\documentclass{article}
\usepackage{ijcai13}

\usepackage{times}

\title{Towards A Theory-Of-Mind-Inspired Generic Decision-Making Framework}
\author{Mihai Polceanu, C\'{e}dric Buche \\
ENI Brest, UMR 6285, Lab-STICC,\\ F-29200 Brest, France \\
polceanu@enib.fr, buche@enib.fr}

\usepackage{graphicx}
\graphicspath{{./images/}}

\begin{document}

\maketitle

\begin{abstract}
Simulation is widely used to make model-based predictions, but few approaches have attempted this technique in dynamic physical environments of medium to high complexity or in general contexts. After an introduction to the cognitive science concepts from which this work is inspired and the current development in the use of simulation as a decision-making technique, we propose a generic framework based on theory of mind, which allows an agent to reason and perform actions using multiple simulations of automatically created or externally inputted models of the perceived environment. A description of a partial implementation is given, which aims to solve a popular game within the IJCAI2013 AIBirds contest. Results of our approach are presented, in comparison with the competition benchmark. Finally, future developments regarding the framework are discussed.
\end{abstract}

\section{Introduction}

When attempting to make a suitable decision within an environment, an agent must first be able to extract data from it via sensors (either virtual or real devices), then process this information in a way that allows a certain level of inference on what to do next and finally apply, if necessary, a set of changes onto the environment or onto itself that lead to a situation that better fits the agent's goals. Although much research has been conducted over the past years, with a wide range of successful solutions in specific domains of activity, general purpose implementations have not been yet achieved and most approaches require case-based parameterization to function.

In this paper, we propose to model the decision mechanisms of a virtual entity by drawing inspiration from studies in cognitive science. To be more precise, our focus is on the human's anticipation ability and capability to adapt while interacting.

\subsection{Cognitive Approach towards Artificial Intelligence Decision Making}


To affirm that an artificial intelligence system reasons like a human, we are required to know how humans think by acquiring insight of the mechanisms of the mind. There are several ways to achieve this: by introspection, psychological experiments on behavior and brain imaging. This work can help to formalize a theory of mind clear enough to make it possible to be expressed in the form of a computer program. The interdisciplinary field of cognitive science advances in this direction by combining experimental psychology techniques and computer models of computational intelligence. In this context, mental simulation is a theory that appears to be a particularly interesting approach to modelling the mechanisms of decision making \cite{Hesslow_02,Berthoz_97,Berthoz_03}.

\subsection{Theory of Mind and Mental Simulation}

The capability of an individual to assign mental states to oneself and others is known in the literature as Theory of Mind (ToM) \cite{Premack78}. Functionally, possessing a model of others' decision process which takes observations as input allows the individual to determine goals, anticipate actions and infer the causes of certain observed behaviors. The two predominant approaches to how the decision process is represented are the Theory-Theory (TT) \cite{cogprints1194}, which implies a "folk psychology" that is used to reason about others in a detached way, and the Simulation Theory (ST) \cite{Goldman05} which sustains that the individual's own decision mechanism is used for inference (simulation), using pretend input based on observations. Regardless the debate over which of the two constitutes the mechanism of reasoning about other individuals \cite{STvsTT,Fisher06,Apperly08}, it is important to note that a model is required to achieve anticipation functionality and this model must be adaptable to match novel contexts and perceptions.

Evidence of mental simulation has been studied in humans \cite{Buckner07,Decety06} where similar brain regions have been observed to become activated when individuals perform remembering, prospection and theory of mind tasks. Furthermore, research in the field of cognitive science shows that mental simulation appears to serve as a tool to understand the self and others.

Research also extends into how people use this type of simulation to mentally represent mechanical systems \cite{Hegarty2004} and to reason on how these will evolve in various conditions. It was suggested that the prediction ability of this approach can be limited by the perceptual capability of the individual and that, in complex situations, subjects cannot accurately predict the precise motion trajectories, but can evaluate within a reasonable margin of error what the effects would be.

\section{Related Work}

Regarding the focus of research that has recently been conducted, two main directions can be distinguished in simulation-based approaches to artificial decision-making, anticipation and learning.

\subsection{Behavioral mental simulation}

Approaches to mental simulation of behavioral traits branch into two separate, but closely related directions, regarding the nature of the simulation target. The first branch is focused on modelling individuals that are highly similar to the "self", case in which the agent that performs the simulation can use its own decision system to infer knowledge about others. The second, which could be seen as an extension of the first, attempts to extrapolate the models used in the simulation to agents with entirely different capabilities.

\subsubsection{Imitation of kin}

As claimed by ST, imitation holds an important position in intention recognition, action anticipation and behavior learning. Such an approach was implemented in Max T. Mouse \cite{Buchsbaum05}, an animated mouse character which uses its own motor and action representations to interpret the behaviors that it observes from its friend, Morris Mouse.

Placing oneself in the context of another individual who is very similar in terms of desires and capabilities can be achieved by using the proprietary model of action from which goals and intentions can be inferred. As an extrapolation of this approach, new behavior can also be learned by associating known interaction with novel objects. However, it has been shown by several studies in neuroscience that the embodiment of an individual can be altered \cite{Schwartz04} without impairing the ability to complete tasks using the new configuration. Therefore, from the point of view of ST, it is convenient for an entity to simulate similar others using its own behavioral mechanisms, but these mechanisms must also be able to evolve.



\subsubsection{Representation of others}

For the claim of ST to hold in the scenario where an agent simulates other individuals that have different embodiments than itself, the correspondence problem \cite{Dautenhahn02,Alissandrakis02} must be approached in order to map others' characteristics onto the agent's own structure. This technique has been used in robots that play roles in teams and infer intentions from human actions in order to help with the goal fulfilment \cite{Gray05}. Inference is achieved by mapping human movements onto the robot's own skeleton so that, through simulation, the robot can determine the significance of the perceived movements.

Other, more TT-inspired approaches make use of specialized models for their simulation targets. This category is especially researched in combination with physical embodiments, such as conversational robots using 3D objects to model their environment \cite{Roy04}. An implementation of a robot was achieved \cite{Kennedy2008,Kennedy2009}, which extends the ACT-R architecture \cite{Anderson04anintegrated} and uses a form of mental simulation to reproduce the decision-making of team mates. In this approach, authors rely on models of human team players from studies in cognitive science to improve the performance of the automated team mate.

The use of self and world models has become a popular requirement, as the importance of higher cognitive traits has been acknowledged especially when coping with novel environments that cannot be modelled only by extrapolating from previously-learnt knowledge, for example in the case of space exploration robots that are bound to encounter unknown situations in their missions \cite{Huntsberger10}.


\subsection{Environmental mental simulation}
\label{sec:env}
Another use of mental simulation is to predict the consequences of actions on the physical environment. For example, in the case of a domino scenario, the use of mental simulation would allow the agent to anticipate different sequences of falling pieces, depending on which domino was first put in motion and the structure of the setup. This result would require an agent to have a physical model of its environment, including properties like mass, gravity, elasticity and friction. It is interesting to note that when verbally presented with the description of a physical scenario, humans tend to construct an image of how it would visually appear in reality and moreover, their image is influenced by their language comprehension ability \cite{Bergen05}.

The mental image of the environment can be viewed as a medium for integrating perceived information \cite{Cassimatis04} and through this approach, an agent capable of simulation based on such a model can anticipate and perform actions in complex interactions with other agents in the environment or with avatars controlled by humans \cite{buche_13a}.

We argue that anticipation in physical environments and in general scenarios cannot be easily achieved with traditional or specific reasoning systems, and that the use of simulation as an anticipation, learning and decision-making mechanism can greatly improve the applicability of the developed system.


\subsection{Discussion}
"Mental simulation" has been computationally approached as an efficient technique for decision-making for intelligent agents in a wide range of real world problems including the anticipation of intruder behavior \cite{Ustun08}, simulation of cultural transmission \cite{Kluwer96} or public transport optimization \cite{fourie2012using}. However, in most cases, a specialized model is developed by domain experts and then used in parametrized simulations to predict the system's future states. Moreover, for this to function, an extended understanding of the problem is required in order to create the models which are used for simulation.

Implementations based on ST seem to behave very well when placed in a familiar environment with self-similar actors and are even able to extrapolate their behavior to account for differing behaviors. However, mapping problems appear when differences increase between the simulating agent and its peers, and such mappings may not be trivial to find. Therefore, in addition to being able to use itself as a theory of mind tool, the agent must also be able to build models of dissimilar objects or other agents.

In order for the agent to build models of its surroundings, it requires a sensor system and suitable learning techniques. The process of simulating an environment based on a learnt model not only enables anticipation and constitutes the base for decision making, but it also reinforces the model itself by allowing the agent to detect discrepancies between the mental image and the reality. Therefore, the creation and improvement of the models can be self-propelled, through simulation.

Once the agent owns a set of models that describe the environment and a representation of itself, also contained within these models, then it is able to create nested simulations. A simulation may contain the representation of the agent that performs the simulation, where information about how the agent itself would behave in certain contexts can be generated, or how its behavior may affect the others. We refer to the concept of nested simulations as "simulation in the simulation".

In this paper we propose a generic framework to model virtual entities (section \ref{sec:generic}) that are able to use simulation in the simulation to anticipate, learn, take decisions and act in dynamic environments where the agent's actions can be influenced or interrupted by other agents or humans. A partial instance of this framework is illustrated in the Angry Birds application (section \ref{sec:Angry}).


\section{Generic Framework}
\label{sec:generic}

In light of the difficulty of decision-making in complex dynamic virtual environments, we propose a ToM-inspired framework that allows the agent to simulate the behaviors of others and also of itself in different contexts. Our aim is to exploit the advantages of simulation as a mechanism to make predictions but also as a learning tool, by using the results of the simulation and comparing them to the outcomes of the real situation. This way the agent can learn and perfect its models of the reality which it inhabits.

\begin{figure}[htp]
\centerline{\includegraphics[width=0.9\columnwidth]{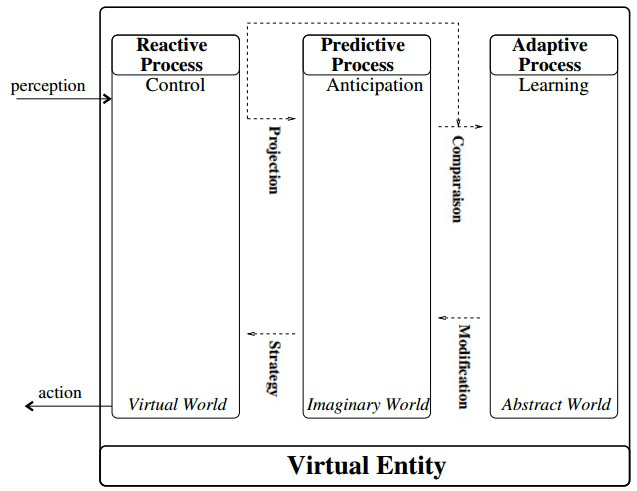}}
\caption{Conceptual overview of the framework}
\label{figure-conceptual}
\end{figure}

Our framework proposal consists in three layers of abstraction which can be regarded as asynchronous processes. The primary level constitutes an interface between the agent and its environment, and consists in all the mechanisms required by the agent to interpret the environment and to act upon it. The imaginary world is a collection of "mental simulations" that the agent uses to anticipate what will happen in the future, based on its current knowledge. These simulations also allow the agent to take decisions in order to achieve a set of goals. The agent's knowledge is represented in the abstract world, which is constantly updated through comparison between anticipations and real events. A conceptual overview of the framework is illustrated in Figure \ref{figure-conceptual}.

In an ideal case, where the agent's models precisely match the environment to be simulated, the problem of predicting what will happen next is solved by simulating the model in advance, at a higher speed. Having computed the natural outcomes of a situation, one can then simulate the effects of agent's actions in the given context. This leads to a tree of possible decisions with associated outcomes, which resolves to a search problem in the space of simulated environment states.

However it so happens that, in most real-world scenarios, the ideal case in which the model is perfectly correct almost never occurs. Therefore, learning algorithms must be used to constantly improve the models. In other words, the correctness of the result depends on how accurate the simulation is, which in turn is dependent of the capability of the model to reflect the rules of the simulated environment.

Given evidence that current learning algorithms are more efficient for certain problems while failing for others, but are however complementary, having a framework that can experiment with multiple learning algorithms would have beneficial influence on the resulted models. Therefore, we propose a heterogeneous architecture of model generation algorithms which can participate in a selection process, based on the correctness of the resulted models regarding evidence from reality. This error evaluation can be done by computing the differences between the state of the real environment and the associated simulated state, at a given time.

While aiming to achieve a completely autonomous agent that can use simulation to create its own models through learning from its environment, it is also practical to allow models developed by domain experts to be integrated, by exploiting the heterogeneous approach of this framework. This way, depending on context, certain parts of the environment can be simulated using human-made models, therefore creating the opportunity to use the framework as a test bed to evaluate the efficiency of automated learning algorithms in given contexts, as well as attempt to improve existing models.

\section{Application: Angry Birds Agent}
\label{sec:Angry}

\begin{figure}[htp]
\centerline{\includegraphics[width=0.9\columnwidth]{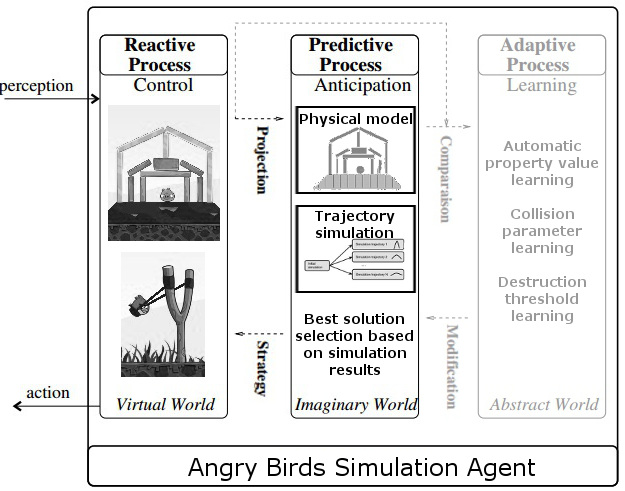}}
\caption{Instantiation of the proposed framework for the Angry Birds bot}
\label{figure-concept-ab}
\end{figure}

In order to evaluate this architecture, we implemented a minimalistic instance of the simulation framework within an agent designed to play the Angry Birds game. Figure \ref{figure-concept-ab} illustrates how the instantiated agent fits into the specifications of the proposed framework, while also mentioning the portion that is to be implemented in a future version (the learning module, drawn in faded color in the figure).

We consider that the AIBirds contest\footnote{AIBirds contest website: http://www.aibirds.org} represents an opportunity to test and evaluate this framework instance in the context of a physical environment with moderately stochastic results which are due to the unavailability of the game's inner mechanisms. In this section we describe the implementation of this agent, with regard to the proposed architecture.

\subsection{Interface with the Game}

In the 2013 edition of the AIBirds contest, a naive agent \cite{ABSoft} was made available to the participants, which contains a set of tools to detect the locations of objects in the Angry Birds scene, and a simple decision algorithm that consists in randomly choosing a pig and shooting the bird on the trajectory that intersects the chosen pig. Object detection is limited to acquiring the bounding boxes, or minimal bounding rectangles (MBR), that are aligned to the game coordinate system.

\begin{figure}[htp]
\centerline{\includegraphics[width=0.9\columnwidth]{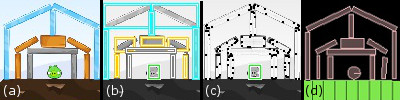}}
\caption{Obtaining correct object positions: (a) original scene image, (b) default bounding boxes, (c) improved algorithm, (d) obtained virtual reconstruction}
\label{figure-boxrot}
\end{figure}

In order for the simulation to reflect the game environment, one must take into account boxes that are rotated either as result of collision or that being their initial state. Our approach to this problem was to improve the existing algorithm by taking into account the already calculated positions for each object, computing the set of pixels that make up the object using a filling algorithm, and fitting the minimum rectangle onto the convex hull of each pixel set by iterating over all edges of the hull. Results of this approach are illustrated in Figure \ref{figure-boxrot}, where (c) represents an overlay of the detected pixels (white) and the points of the convex hull (black), and (d) shows the minimum rectangles fitted on the convex hulls, which also represent the simulation objects (image taken from a rendering of our virtual scene).

Additional to this detection process, the dimensions of detected rectangles are equalized with similar instances, based on the fact that the game uses fixed size objects.

Furthermore, because it is possible for the reconstruction to be unstable, objects may fall without being touched. To avoid such scenarios, the agent makes the assumption that the configuration in which the environment is found before a shot is a stable one. Therefore, objects are held in place within the simulation, and must become "active" in order to move. At the start of the simulation, only the shooting bird is active, then it activates other objects upon collision. In other words, each object becomes active if a collision happened between itself and an already active object.

Another approach \cite{Jochen13} to extracting general solid rectangles (GSR) from MBRs consists in defining an algebra to describe the GSRs and computing all possible contact combinations and finding a stable configuration that satisfies all the requirements. Unlike our approach, it is interesting to note that using an algebra to reason directly on MBRs does not require further analysis of the original image. However, due to the assumption that the MBRs are correct, the case in which one object is not detected can trigger a significantly different result, as it may have played an important role in the scene configuration. Furthermore, in the rare situation where MBRs are vertically aligned, there may exist two mirrored possibilities to represent the GSRs.

Using original image analysis to compute GSRs is arguably more computationally expensive, but it allows to limit detection errors to a local scope. Moreover, by using a convex hull to determine the rectangles, the number of points of the hull can be used to decide, with an acceptable accuracy, which objects are rectangular and which are circular (i.e. circular objects have more hull vertices than rectangular objects).

\subsection{Knowledge Representation}

Once detected, objects are mapped onto the agent's knowledge consisting of a set of classes that represent physical objects and materials. All objects in the simulation are animated by a two-dimensional physics engine, which corresponds to a domain expert model integration within the proposed framework. The knowledge representation is inspired from the object-oriented paradigm, thus offering the possibility of modelling, among others, attributes and inheritance. Each type of bird (red, yellow, etc.) inherits the main Bird class behavior, which in turn is a physical Object. The goal of this representation is to be able to accurately describe the effects of performed actions and the environment evolution in time.

This representation allows different behaviors to be easily assigned to sets of instances. In the case of this application, the values of attributes are empirical in origin and not automatically learned, for convenience. However, this will be the ultimate goal of the proposed framework.

\subsection{Simulation in the Simulation}

\begin{figure}[htp]
\centerline{\includegraphics[width=0.9\columnwidth]{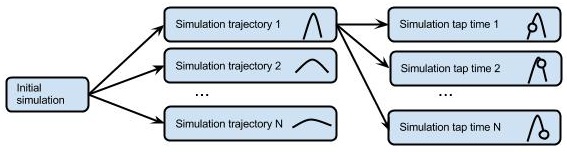}}
\caption{Simulation tree with two levels: trajectories and tap times}
\label{figure-simtap}
\end{figure}

Once the environment structure is interpreted, the agent will use its imaginary world to create a number of simulations that differ by a small shooting angle (Figure \ref{figure-simtap}).

The depth of the simulation tree can be increased, by adding the "tap" function which triggers special behavior in some types of birds. For each shooting angle, the agent is able to choose a time to perform the tap which consists in another array of possibilities, therefore adding another level in the simulation tree (Figure \ref{figure-simtap}). This is done similarly to the first level, through simulation duplication. This method proves to be computationally inexpensive, as the initial object recognition and mapping are not remade, but their results are copied and simulated in another way (eg. different angles or tap times, but with the same initial scene configuration). In the simulations, a special bird behavior triggers similar effects as in the real game, for example blue birds spawn new instances at different angles, yellow birds gain a speed boost, black birds explode and white birds shoot projectiles downwards. This way, the model can be modified to better fit the game without changing the decision making process.

Based on the results of the simulations, the agent will choose one which best suits its goal: to destroy pigs. However, the mere evaluation of how many pigs are killed in the simulation does not prove to give the most successful results. This issue is not because the selection process is incorrect, but because the model of the environment is not completely accurate. This causes the simulations to give similar but not precise outcomes compared to the actual game scenario. For example, it may happen that in one simulation, conditions are just right for a record number of pigs to be killed, although it is not the same case in the real game; this leads to a erroneous score record, and therefore this simulation is incorrectly chosen as the final decision. 

Coping with uncertainty caused by a partially correct model can be addressed by considering in advance that results are stochastic in nature, and therefore taking into account a range of results instead of isolated cases. In other words, instead of evaluating each simulation by its final result, one can evaluate a set of results from minimally different simulations to obtain a probabilistically good decision. This way, isolated cases of false success can be pruned out by the selection algorithm, leaving higher chances for the agent to take a truly efficient decision.

\subsection{Preliminary Results}

Our implementation currently creates a number of 106 simulations which differ by a shooting angle of 0.01 radians from each other. Normal simulation time is approximately 15 seconds, however the simulations are executed at three times the normal speed, due to the time limitation imposed by the contest rules.

We compare our implementation with the default "naive" agent provided by the competition organizers. This choice for benchmark was made due to the fact that the current default agent is an extension of the winning agent of the 2012 edition of the contest, that provides an improvement of the visual recognition module.

Because the default agent is based on a random target selection, and this may influence the resulted scores, we limited the levels used in our comparison to the first 9 levels of the freely available "Poached Eggs" episode of Angry Birds, which only feature red birds that do not have any special ability.

\begin{table}[h]
\begin{center}
\begin{tabular}{|c|c|c|c|c|c|}
\hline
\multicolumn{1}{|c|}{{\bf Lvl}} & \multicolumn{1}{|c|}{{\bf{Trial 1}}} & \multicolumn{1}{|c|}{{\bf{Trial 2}}} & \multicolumn{1}{|c|}{{\bf{Trial 3}}} & \multicolumn{1}{|c|}{{\bf{Trial 4}}} & \multicolumn{1}{|c|}{{\bf{Avg}}} \\ \hline
1       & 30090 & 28960 & 28970 & 28970 & \emph{29247.5} \\ \hline
2       & 42650 & 34160 & 34160 & 43180 & \emph{38540.0} \\ \hline
3       & 31670 & 40730 & 40260 & 40260 & \emph{38230.0} \\ \hline
4       & 28470 & 18870 & 28160 & 28330 & \emph{25957.5} \\ \hline
5       & 65140 & 64500 & 64710 & 63510 & \emph{64465.0} \\ \hline
6       & 34010 & 24450 & 25630 & 24490 & \emph{27145.0} \\ \hline
7       & 20020 & 29390 & 27680 & 22910 & \emph{25000.0} \\ \hline
8       & 38280 & 36170 & 58010 & 38650 & \emph{42777.5} \\ \hline
9       & 32870 & 31770 & 32930 & 29090 & \emph{31665.0} \\ \hline

\bf{tot.} &\bf{323200}&\bf{309000}&\bf{340520}&\bf{319390}&\bf{\emph{323027.5}} \\ \hline
\end{tabular}
\caption{Scores of the default naive agent}
\label{table-scores-naive}
\end{center}
\end{table}

After evaluating the naive agent that was provided by the competition organizers (Table \ref{table-scores-naive}), we found it to fail a total of 22 times while playing 4 trials of 9 levels each. In comparison, the scores of the agent proposed in this paper are shown in Table \ref{table-scores-sim} and exhibit a 10.9\% average improvement over the naive agent, and only 4 total failures to finish a level.

\begin{table}[h]
\begin{center}
\begin{tabular}{|c|c|c|c|c|c|}
\hline
\multicolumn{1}{|c|}{{\bf Lvl}} & \multicolumn{1}{|c|}{{\bf{Trial 1}}} & \multicolumn{1}{|c|}{{\bf{Trial 2}}} & \multicolumn{1}{|c|}{{\bf{Trial 3}}} & \multicolumn{1}{|c|}{{\bf{Trial 4}}} & \multicolumn{1}{|c|}{{\bf{Avg}}} \\ \hline
1       & 32350 & 32350 & 32330 & 32330 & \emph{32340.0} \\ \hline
2       & 60820 & 52070 & 60630 & 51530 & \emph{56262.5} \\ \hline
3       & 42630 & 42630 & 41820 & 42630 & \emph{42427.5} \\ \hline
4       & 28120 & 28120 & 28430 & 28120 & \emph{28197.5} \\ \hline
5       & 64330 & 69100 & 63520 & 58070 & \emph{63755.0} \\ \hline
6       & 24470 & 33510 & 26270 & 33510 & \emph{29440.0} \\ \hline
7       & 33200 & 22660 & 23300 & 31940 & \emph{27775.0} \\ \hline
8       & 34710 & 47520 & 39120 & 44820 & \emph{41542.5} \\ \hline
9       & 24370 & 40010 & 40920 & 41440 & \emph{36685.0} \\ \hline

\bf{tot.} &\bf{345000}&\bf{367970}&\bf{356340}&\bf{364390}&\bf{\emph{358425}} \\ \hline
\end{tabular}
\caption{Scores of the simulation in the simulation agent}
\label{table-scores-sim}
\end{center}
\end{table}

The most failures of the naive agent were on the 7th level (10 failures) and the second most on the 4th level (4 failures). Our implementation also experienced difficulty in level 7 but with only 3 failures, and on level 9 with only one failure.

During the trials, the situation occurred where our agent obtained the same score on two or more attempts. Identical scores are due to the same decision taken more than once, as the visual recognition of the objects in the scene returned the same results. It is important to note that most of these situations occurred when the agent solved the level using one bird (levels 1, 3 and 4). An exception to this rule was level 6, where due to the lower number of objects, the scene tends to give the same results after the first shot. However, this shows that given exactly the same conditions, the simulations are stable enough to output the same decision for the agent.

\begin{figure}[htp]
\centerline{\includegraphics[width=0.9\columnwidth]{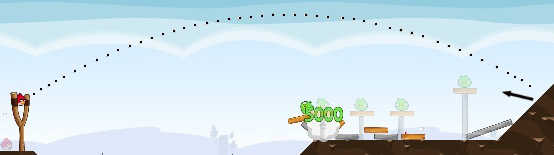}}
\caption{Best solution for Level 2 of "Poached Eggs" episode, found by our implementation (overlaid original state and highlighted trajectory)}
\label{figure-bestshot}
\end{figure}

Tests show that errors in the model are bound to make the agent take less efficient decisions, because the simulations do not correctly reflect the game reality. However, when the reconstruction has a higher level of correctness, the agent is able to find the most effective strategy to complete the game, as shown in Figure \ref{figure-bestshot} which illustrates the second level of the freely available "Poached Eggs" episode of Angry Birds.

These results indicate that it is possible to take good decisions even if the model of the environment does not perfectly match the game environment. However, to achieve such good results in the majority of scenes, further efforts must be made to improve the model.

\section{Conclusions and Future Work}

In this paper we have described a generic framework that addresses the problem of decision-making, learning and anticipation in dynamic environments. Our approach is motivated by the need of a non-specific framework that can be used in multiple contexts and that is open to heterogeneous models to be inputted or learned, in order to maximize the potential of existing techniques for problem solving.

The architecture of the proposed framework consists in three main components that can be executed in parallel and revolve around the concept of simulating events in advance, which allows not only for anticipating results, but also for learning and improving the models based on the differences observed between the simulations and the real environment. We also consider that employing multiple models to describe the entities that populate the environment will build the basis of a generic selection system of most suited models, which would lead to a continuously improving performance of the system.

A version of the current implementation will be submitted to the IJCAI2013 AIBirds contest to evaluate the efficiency of this approach in this context. For this, we will focus especially on the game interface module and the special bird abilities. This implementation only performs environment mental simulation (section \ref{sec:env}).

As future development of the proposed architecture, we intend to make use of a knowledge and procedural behavior representation framework such as MASCARET \cite{Querrec}, which would provide a suitable meta-model for our framework. Using this approach, structural and behavioral models can be inputted by a domain expert or automatically generated from within the framework. Our goal is to obtain an agent that is able to reason and anticipate complex situations such as those that occur in virtual environments for training \cite{querrec_03a}. The general-purpose of the solution will be further tested using various environments, including the attempt to exit the virtual via a robotic embodiment, although this implies additional interface-related challenges.

Furthermore, as one of the drawbacks of our current instance is that it does not contain automatic learning capability (i.e. the Abstract World layer of the framework), we intend to develop such functionality in an extended implementation of the proposed framework.

We consider that one of the most important challenges in this approach stems from the difficulty of correctly modelling the real environment and therefore through automatic learning (automatic model amelioration), differences between the model and the reality will be diminished, leading to an improvement in decision efficiency.

\bibliographystyle{named}
\bibliography{ijcai13}

\end{document}